\documentclass[acmlarge]{acmart}
\usepackage{graphics}
\usepackage{hyperref}
\usepackage{url}
\usepackage{microtype}      
\usepackage{color}
\usepackage{wrapfig}
\usepackage{graphicx}
\usepackage{epstopdf}
\usepackage{cuted}
\usepackage{bbm}
\usepackage{multirow}
\usepackage{makecell}
\usepackage{multirow}
\usepackage{soul}
\usepackage{arydshln}
\usepackage{algpseudocode}
\usepackage{algorithm}
\usepackage{flexisym}

\usepackage{booktabs}
\usepackage{xspace}
\usepackage{amsmath} 
\usepackage{mdframed}
\usepackage{enumitem}
\usepackage{pifont}
\usepackage{adjustbox}

\AtBeginDocument{%
  \providecommand\BibTeX{{%
    \normalfont B\kern-0.5em{\scshape i\kern-0.25em b}\kern-0.8em\TeX}}}

\acmJournal{POMACS}
\acmVolume{37}
\acmNumber{4}
\acmArticle{111}
\acmMonth{8}

\setcopyright{rightsretained} 
\acmJournal{IMWUT}
\acmYear{2022} \acmVolume{6} \acmNumber{1} \acmArticle{24} \acmMonth{3} \acmPrice{}\acmDOI{10.1145/3517225}

\begin{document}

\newcommand{\xin}[1]{\textcolor{red}{\emph{{Xin:}~{#1}}}}
\newcommand{\yuntao}[1]{\textcolor{red}{\emph{{Yuntao:}~{#1}}}}
\newcommand{\dan}[1]{\textcolor{blue}{\emph{{Daniel:}~{#1}}}}
\newcommand{\sinan}[1]{\textcolor{brown}{\emph{{}~{#1}}}}

\newcommand*{\projectname}{MobilePhys\xspace}

\newcommand{\revision}[1]{\textcolor{black}{#1}} 
\newcommand{\reviewcomment}[1]{\vspace{0.0cm}\begin{mdframed}[backgroundcolor=gray!20]#1\end{mdframed}\vspace{0.4cm}}
\newcommand{\xmark}{\ding{55}}%
\newcommand{\cmark}{\ding{51}}%
\makeatletter
\def\blfootnote{\xdef\@thefnmark{}\@footnotetext}
\makeatother

\title{MobilePhys: Personalized Mobile Camera-Based Contactless Physiological Sensing}


\author{Xin Liu * \ddag}
\email{xliu0@cs.washington.edu}
\affiliation{%
  \institution{University of Washington}
  \city{Seattle}
  \state{WA}
  \country{USA}
}

\author{Yuntao Wang *}
\affiliation{%
  \institution{Tsinghua University}
  \city{Beijing}
  \country{China}
}

\author{Sinan Xie *}
\affiliation{%
  \institution{Tsinghua University}
  \city{Beijing}
  \country{China}
}

\author{Xiaoyu Zhang}
\affiliation{%
  \institution{Tsinghua University}
  \city{Beijing}
  \country{China}
}

\author{Zixian Ma}
\affiliation{%
  \institution{Zhejiang University}
  \city{HangZhou}
  \country{China}
}

\author{Daniel McDuff}
\email{damcduff@microsoft.com}
\affiliation{%
  \institution{Microsoft Research}
  \city{Redmond}
    \state{WA}
  \country{USA}
}

\author{Shwetak Patel}
\email{shwetak@cs.washington.edu}
\affiliation{%
  \institution{University of Washington}
  \city{Seattle}
  \state{WA}
  \country{USA}
}

\renewcommand{\shortauthors}{Liu and Wang and Xie, et al.}
\blfootnote{* Equal Contribution}
\blfootnote{\ddag	Lead Student Author}
\begin{abstract}

Camera-based contactless photoplethysmography refers to a set of popular techniques for contactless physiological measurement. The current state-of-the-art neural models are typically trained in a supervised manner using videos accompanied by gold standard physiological measurements. However, they often generalize poorly out-of-domain examples (i.e., videos that are unlike those in the training set). Personalizing models can help improve model generalizability, but many personalization techniques still require some gold standard data. To help alleviate this dependency, in this paper, we present a novel mobile sensing system called \projectname, the first mobile personalized remote physiological sensing system,  that leverages both front and rear cameras on a smartphone to generate high-quality self-supervised labels for training personalized contactless camera-based PPG models. To evaluate the robustness of \projectname, we conducted a user study with 39 participants who completed a set of tasks under different mobile devices, lighting conditions/intensities, motion tasks, and skin types. Our results show that \projectname significantly outperforms the state-of-the-art on-device supervised training and few-shot adaptation methods. Through extensive user studies, we further examine how does \projectname perform in complex real-world settings. We envision that calibrated or personalized camera-based contactless PPG models generated from our proposed dual-camera mobile sensing system will open the door for numerous future applications such as smart mirrors, fitness and mobile health applications.
\end{abstract}
\maketitle
\section{Introduction}

One of the visions of ubiquitous computing is the ability for people to interact with computing using any device. Today, some of the most ubiquitously available sensors are RGB cameras. Camera-based contactless physiological sensing refers to a set of techniques that enable contactless measurement of cardio-pulmonary signals and their related vitals signs, such as heart rate, respiration rate and blood oxygen saturation. Unobtrusive physiological sensing technology could help advance the vision of ubiquitous computing in numerous contexts, but perhaps most directly in health, well-being and affective computing applications. Cardiac and respiratory processes change the appearance of the body in several ways.
Camera-based contactless photoplethysmography or remote photoplethysmography (rPPG) involves the measurement of very subtle changes in  light reflected from the skin to capture the photoplethysmogram. When the light hits the skin, the amount that is absorbed is influenced by the current peripheral blood volume. Subtle motions caused by blood pumping around the body can also be measured using optical flow patterns to recover the ballistocardiogram (BCG)~\cite{balakrishnan2013detecting}. The resulting pulse waveforms (PPG and BCG) can be used to derive heart rate and heart rate variability~\cite{poh2010advancements}. Based on natural sinus rhythm, the pulse signal can also be used to estimate the respiration rate~\cite{poh2010advancements}. However, pulmonary signals are often more obvious based on the motion of the torso due to the physical motion of inhaling and exhaling. A combination of mechanical and optical information typically would provide the richest signal. In this paper, we focus on camera-based contactless PPG measurement.

The COVID-19 pandemic acutely highlighted the importance and utility of camera-based contactless physiological sensing technology~\cite{song2020role,smith2020telehealth}. The desire to protect healthcare workers and patients and reduce the need for people to travel illustrates how ubiquitous sensing could be used at scale. However, most people still do not have a way to measure the necessary signals at home.  In this paper, we contribute towards this goal by proposing a system to easily personalize contactless PPG measurement models using a smartphone. These customized and personalized models produced from our system enable any device equipped with an RGB camera (e.g., a smart mirror, fitness equipment, such as Peloton or Mirror or even possibly a baby monitor), to provide comfortable, in-situ vital monitoring. Compared to traditional pulse oximeters using contact PPG techniques, camera-based contactless physiological sensing also provides a unique advantage to reduce the risk of infection for vulnerable patients and discomfort caused by obstructive wires~\cite{villarroel2019non}.

Although camera-based contactless physiological sensing comes with many advantages, it also presents various technical challenges. First, there is still an accuracy gap between contact sensors and camera-based contactless solutions. The US Federal Drug Administration (FDA) requires a new device for cardiac monitoring to have substantial equivalence in accuracy with the FDA-approved devices. Unfortunately, none of the camera-based contactless systems has reached the bar of requirements of FDA-approved cardiac devices. Second, current camera-based contactless physiological sensing systems are especially sensitive to numerous noises such as lighting conditions and motions from different activities. Prior research has shown that the accuracy of these systems is significantly reduced while introducing such noises \citep{chen2018deepphys,liu2020multi}. Third, there is a large individual difference in appearance (e.g., gender, skin type, makeup, hair) and physiology (e.g., blood volume dynamics). Creating a generalizable system for these conditions presents an interesting challenge.

One way to solve the aforementioned limitations is to train a supervised model with a large and diverse set of training data that contains samples that exhibit the types of variations expected at test time (e.g., race, lighting, motion). However, collecting such a large and high-quality physiological dataset is challenging. Although the process of data collection requires significant resources for recruitment and management, it is also risky to disclose the sensitive identity and physiological information of participants. Hence, traditional supervised training using a large-scale dataset is laborious and difficult for building an unbiased and generalizable camera-based contactless physiological sensing system.

In traditional clinical settings, physicians often use high-end medical devices to help calibrate customer-level medical sensors for each patient. The procedure of calibration helps combat individual differences in sensor performance and strengthens the validity of the output. Therefore, training a personalized model for each individual in different environments is ideal. However, getting high-quality synchronized video and ground-truth physiological signals for training a personalized model is difficult. This is especially complicated if patients want to calibrate with their smartphones' cameras because external medical sensors are barely compatible with smartphones. A mobile system that performs self-calibration in camera-based contactless physiological sensing is attractive.  

Meta learning is an emerging technique in machine learning that aims to learn how to learn a task faster~\cite{hospedales2020meta}. The goal of meta learning is to learn a quick learner for a new task (e.g., person). However, most meta learning tasks assume that ground-truth labels are available in the dataset during training which indeed is not the case for most applications. Because of the recent advancement of mobile sensors in smartphones, smartwatches, and IoT devices, mobile sensing systems now have the ability to provide some high-fidelity sensor data, even ground-truth labels for some applications. We believe the interplay between meta learning and mobile sensing systems has been underused in many domains.

In this work, we use contactless PPG measurement as an example to demonstrate how novel sensing systems can provide reliable pseudo physiological labels to meta learning algorithms for training few-shots adaption models. We propose a self-calibrating meta-learning system called \projectname, which leverages both front and rear cameras available on smartphones and the ability for us to measure physiological signals from multiple parts of the body. Specifically, we design a system that simultaneously measures the PPG signal from the finger-tip and the face of the subject during a calibration period to personalize a contactless measurement model that only relies on analyzing the face. The pseudo PPG labels generated from \projectname using the rear camera and the index finger can provide similar waveform quality to ground-truth PPG signals from medical-grade pulse oximeters. We demonstrate that this is also reliable in challenging real-world conditions (e.g., motion, lighting and darker skin types). Models customized or personalized using \projectname could then be deployed on the phone or shared with other smart devices (such as a laptop or smart mirror~\cite{poh2011medical}) to enable convenient contactless measurement.

In summary, we propose a novel smartphone-based personalized physiological sensing system that leverages the back and front RGB camera to perform self-adaptation. More specifically, our contributions include:

\begin{itemize}
    \item Proposing a novel mobile dual camera-based contactless physiological sensing system that generates high-quality pseudo PPG labels for few-shot personalization and adaptation.
    \item Demonstrating that we can leverage contact finger-tip PPG signal derived from the smartphone's rear camera to train a personalized camera-based contactless physiological neural network. 
    \item Exploring and evaluating the performance of \projectname under different conditions such as different mobile devices, lighting conditions, motions, activities, skin types, and camera settings through comprehensive user studies.
    \item Studying and investigating mobile camera settings, which we believe will be valuable for guiding future research in mobile physiological sensing.
    \item Finally, we collected and will release the first-ever multi-modality mobile camera-based contactless physiological dataset with different mobile devices, lighting conditions, motions, activities, and skin types. The documented dataset has gold standard oximeter recordings and synchronized finger PPG signals from the rear camera, and face videos along with other sensor signals (e.g., IMU, ambient light, etc.) from the smartphone. The dataset comprises close to six hours of video and sensor data from 39 participants. 
\end{itemize}

\section{Related Work}

\subsection{Mobile and Camera-based Contactless Physiological Sensing}

Thanks to their ubiquity and portability, smartphones are becoming a popular tool for monitoring activity and health parameters. Smartphones cameras can be used to effectively measure the PPG~\cite{li2019current} using imaging or camera-based contactless PPG~\cite{mcduff2015survey}. There are two primary forms this can take: 1) contact measurement in which the subject places the camera against their skin (usually by placing their finger-tip over the camera lens) and the flash is optionally used to illuminate their skin; 2) remote measurement in which the subject faces the camera and a region of skin (usually on the face) is segmented and analyzed using ambient illumination. Contact measurement is typically the most robust and PPG and oxygen saturation (SpO$_{2}$) measurements~\cite{scully2011physiological} are well established. Research is continuing into how these techniques can be used to measure blood pressure~\cite{schoettker2020blood}.

However, contact measurement is often not convenient. Camera-based contactless measurement can be used for opportunistic measurement when people unlock their smartphone using the front camera. Moreover, due to the simplicity and portability of ubiquitous cameras, camera-based contactless measurement also has a great potential to provide scalable and accessible health sensing. While there any attractive properties of non-contact systems including comfort, scalability, and convenience, it still has numerous challenges involved in accurately measuring physiological signals such as PPG signal, as the distance between the camera and the region of interests adds greater lighting variations, possibilities for objects to occlude the skin and/or there be motions of the ROI relative to the camera.

Imager or camera-based contactless physiological sensing can be performed using handcrafted signal processing methods or supervised learning. Fundamentally, these approaches are based on optical models that offer a way to model the interaction between ambient illumination and skin.  The Lambert-Beer law (LBL) and Shafer’s dichromatic reflection Model (DRM) have been employed as inspiration for these algorithms. Thanks to their simplicity, signal processing-based methods were the first to be used in pulse measurement ~\cite{poh2010advancements,6523142,7565547} and respiration measurement~\cite{Tarassenko_2014}. These pipelines primarily use color space conversions and signal decomposition. Early work ~\cite{Li_2014_CVPR,Verkruysse:08} only used green channel data as the PPG signal strength is typically strongest in the corresponding frequency bands. Subsequent research established that combining multiple color channels leads to a solution that is more robust to variation in environment (e.g., ambient lighting) and motion~\cite{de_Haan_2014,6523142}. Most of the previous work employed Principal Component Analysis (PCA)~\cite{6894148} or Independent Component Analysis (ICA)~\cite{Poh:10,poh2010advancements,Mcduff_1improvementsin} for signal decomposition. However, these approaches are susceptible to noise from head movements and variations in lighting conditions. More advanced approaches solved this issue by taking advantage of patients' skin characteristics knowledge ~\cite{6523142,7565547} giving somewhat more robust results. However, it is still difficult for these handcrafted signal processing pipelines to successfully separate the physiological signals from other pixel variations, many of which may be much larger than the subtle changes resulting from physiological processes. 

Supervised learning  methods often achieve superior performance compared to unsupervised signal processing approaches. These methods are able to capture highly non-linear relationships between the physiological signal and facial videos. DeepPhys~\cite{chen2018deepphys} was the first end-to-end neural approach for camera-based contactless physiological measurement. The model learns a soft-attention mask learning appearance information related to the physiological signals. The attention mechanism helps reduce noise by adapting the region of interest (ROI). Subsequent research has succeeded in leveraging neural networks for BVP or respiration measurement\citep{Spetlik2018VisualHR, yu2019remote, 9050507, 8879658}. For instance, RhythmNet~\cite{8879658} computes spatial-temporal maps of the ROI facial areas to represent the HR signal passed to the succeeding neural network. Song et al.~\cite{9050507} proposed transfer-learning a convolutional neural network (CNN). The model is first trained using spatio-temporal images generated using synthetic contactless PPG signals, and then real videos are used to refine the model. However, one drawback of these methods is the computational overhead of resulting networks. Given the ubiquity of RGB cameras on smartphones, ideally, a solution would run on these types of devices. Multi-task temporal shift attention network was proposed as one solution to enable on-device camera-based contactless cardiopulmonary monitoring on a smartphone~\cite{liu2020multi} and reached an inference rate of over 150 frames per second. 

However, neural networks still face various challenges. Individual differences in appearance (e.g., skin type, with glasses or not, pulse dynamics), environmental variations (e.g., light intensity, spectral composition) and motion  (e.g., talking, head movement) make it hard to train an algorithm that generalizes well to unseen data. Camera settings and hardware sensitivity also vary significantly among devices, and prior research has shown that video compression also affects physiological measurement result~\cite{mcduff2017the}. As a result, supervised models often perform significantly worse on cross-dataset evaluation than within-dataset evaluation. 

\subsection{Meta-Learning and Personalized Physiological Sensing}

Learning from a small number of samples or observations is a hallmark of an intelligent agent. However, traditional machine learning systems do not perform well under such constraints. Meta-learning approaches tackle this problem, creating a general learner that is able to adapt to a new task with a small number of samples \citep{hospedales2020meta}. Previous work in meta-learning had focused on supervised computer vision tasks \citep{zoph2018learning, snell2017prototypical} and applied these methods to image analysis~\citep{vinyals2016matching, li2017meta}. In the video domain, meta-learning has been successfully applied in object and face tracking~\citep{choi2019deep,park2018meta}. In these tasks, the learner needs to adapt to the individual differences in the appearance of the target and then track it across frames, even if the appearance changes considerably over time in the video. Choi et al. \cite{choi2019deep} present a matching network architecture providing the meta-learner with information in the form of loss gradients obtained using the training samples. 

\begin{table}[ht]
\begin{center}
 \caption{Comparison of State-of-the-Art Methods in Camera-Based Contactless Physiological Sensing}
 \begin{tabular}{c  c  c  c} 
 \toprule
 \textbf{Method} & \textbf{On-Device} &  \textbf{Adaptation}  & \textbf{Reliable Pseudo Label} \\
 \toprule
TS-CAN \cite{liu2020multi} & \cmark & \xmark & \xmark \\
Meta-rPPG \cite{lee2020meta} & \xmark & \cmark & \xmark \\
MetaPhys \cite{liu2021metaphys} & \cmark & \cmark & \xmark \\
MobilePhys (Ours) & \cmark & \cmark & \cmark \\
 \hline
 \bottomrule
 
\end{tabular}
\label{table:related_work}
\end{center}
\end{table}

As deep learning methods struggle to generalize to unseen tasks and data, developing a personalized physiological sensing model using only a few unlabeled samples is promising. Encouraged by success on other tasks, we leverage meta-learning as the way of adapting our camera-based contactless PPG sensing algorithms.
This work builds upon two specific examples of meta-learning applied to PPG measurement. Meta-rPPG ~\cite{lee2020meta} first introduced meta-learning for heart rate estimation. It achieves self-supervised weight adjustment by generating synthetic gradient and minimizing prototypical distance. MetaPhys ~\cite{liu2021metaphys} was then proposed is based on Model-Agnostic Meta-Learning (MAML)~\cite{finn2017model}. It took advantage of the advanced on-device network architecture ~\cite{liu2020multi} and probed into both supervised and unsupervised training regimes, both of which yielded satisfactory results. For supervised learning, ground-truth signal comes from medical-grade contact sensors, while in the unsupervised version, pseudo labels are used instead in the meta-learner training process. Though effective, these prior works rely much on synchronized video and ground truth obtained from medical-grade devices. However, it is difficult and laborious to collect a large-scale physiological dataset. In this work, we propose a mobile sensing system that leverages both front and rear cameras to generate contact PPG labels and personalize a camera-based contactless physiological system to address this issue. We summarize the difference between popular recently published neural methods in camera-based contactless physiological sensing in Table \ref{table:related_work}. Since our goal is to develop an on-device mobile personalization system for camera-based contactless physiological sensing, \projectname has clear benefits over the state-of-the-art methods. It is also the only system that can help generate reliable pseudo labels under various contexts (e.g., motion, lighting, skin types).

\section{Method}
In the following sections, we first describe the design of our sensing system for personalized camera-based contactless physiological sensing called \projectname. We then explore how to combine meta learning with \projectname to perform few-shot personalization without the need for clinical-grade sensors.

\begin{figure*}[t!]
  \centering
  \includegraphics[width=1\textwidth]{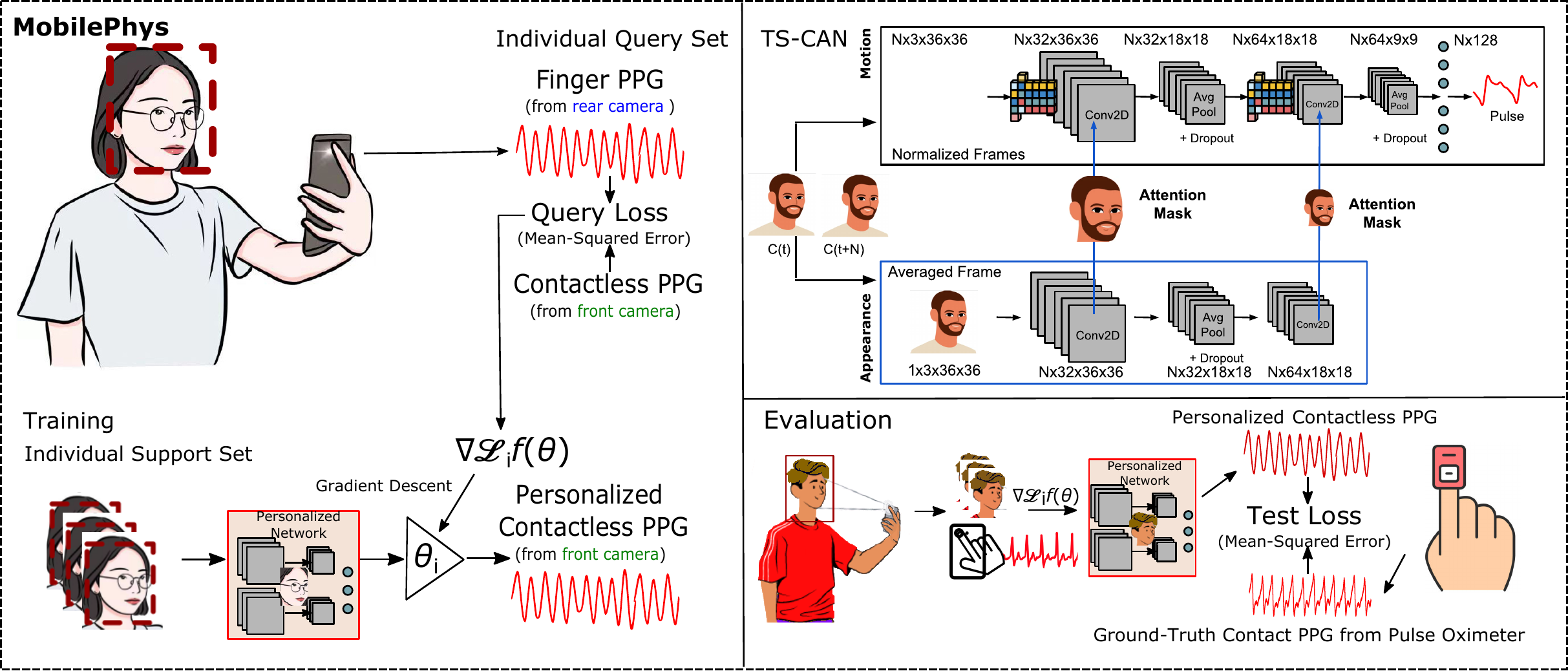}
  \caption{We present \projectname, a novel mobile camera-based contactless physiological sensing system that leverages rear camera to generate self-supervised "ground-truth" PPG label to help train a contactless and personalized physiological model}
  \label{fig:mobilephys}
\end{figure*}

\subsection{Mobile Camera-based Contactless Physiological Sensing System}

We propose a novel mobile dual-camera sensing system to perform personalized physiological measurement called \projectname (Figure \ref{fig:mobilephys}). We leverage the fact that modern smartphones are equipped with at least two RGB cameras. The one on the front of the device is typically used for ``FaceTiming'' and selfies, and another one on the back of the device is used for photography and video recording. Thanks to significant investment in camera technology, these mobile cameras are typically high-quality optical sensors (e.g., low signal-to-noise ratio and high spatial resolution).
Previous work has shown that the PPG signal and heart rate can be measured accurately from the fingertip and face using these cameras~\cite{coppetti2017accuracy}. However, there are no examples that combine these sensors together to create a self-supervised system.

The basic principle behind both contact and non-contact PPG is that the volume of blood flow influences light absorption in the skin. The optimal frequency is approximately 540nm (where the absorption of light by hemoglobin is highest)~\cite{blackford2018remote}. Thus customized PPG sensors are usually designed with green light sensors (e.g., Apple Watch). Unfortunately, most smartphones are not equipped with green light sensors specifically designed for PPG measurement. However, the camera can be used as a proxy.

Our smartphone-based sensing system leverages the rear camera to provide a reference ``ground-truth'' PPG signal for personalizing a non-contact algorithm that measures the PPG directly from a video of the face. More specially, for a short calibration recording, we spatially average the red channel camera frames from the backward-facing (rear) camera while the participant has their finger over it.  We turn on the built-in flashlight to increase the illumination. Simultaneously we capture frames of the participant's face using the front RGB camera. As ~\ref{fig:mobilephys} demonstrates, a person holds the smartphone while pressing their index finger on the rear camera. By using this dual-camera sensing system, it gives us perfectly synchronized contact and non-contact PPG data. Through using a few-shot meta learning algorithm (described in the next section), we then train a personalized non-contact PPG measurement algorithm (bootstrapping from a pretrained network). Each user only needs to hold the smartphone in this way for 18 seconds to create their own personalized physiological sensing model and, in the future can leverage that model without needing to place their finger on the rear camera.

\subsection{Design and Implementation}

While it may seem a simple concept, implementing \projectname was not trivial. We used a Xiaomi MI 8 for this project because it has a well-supported API for dual-camera development. To validate our method, we created a system to synchronize gold standard PPG measurements from a clinical-grade pulse oximeter, fingertip video recordings from the rear camera, and facial video from the front camera. Something that we observed was that many smartphone cameras struggled to maintain a consistent sampling rate while recording from both cameras and connecting a medical-grade pulse oximeter. This may in part be due to power consumption. To solve this limitation, we developed a customized router-based flask system in which we can connect the medical-grade pulse oximeter to a laptop while synchronizing the external pulse oximeter and the data generated from our mobile app. More specifically, our smartphone app system provides a trigger signal to the laptop that is connected with the gold standard pulse oximeter when the mobile app starts recording. 
The laptop and the mobile phone were connected to a local network to minimize delay in communication. We viewed precise synchronization as an important part of our system so that the data can ultimately be used for training the camera-based contactless measurement algorithm that predicts the pulse waveform value for every video frame. 

In building our system, we identified several important parameters that impacted the data quality. Smartphone video recording software has many automatic controls to make video recordings more visually pleasing. These include but are not limited to automatic exposure, white balance and sensitivity controls. We found that these controls on smartphones were more aggressive than those on other video capturing devices (e.g., DSLRs). This is perhaps because these devices are expected to operate with little user control in many contexts and lighting conditions. Therefore, as part of our analysis, we explored how these mobile camera controls affected the accuracy of camera-based contactless PPG measurement. Since camera-based contactless physiological measurement aims to capture very small changes of color changes from the skin, subtle noise can easily corrupt or interfere with the signal we aim to extract. Surprisingly, this type of systematic analysis is not often performed and we believe that characterizing the impact of these features will help other researchers who are building camera-based contactless physiological sensing systems.

To recover the PPG from the rear camera video, we used the shift method to decode the Android color integer format pixel data into four ARGB components \footnote{https://developer.android.com/reference/android/graphics/Color\#decoding}, where A represents the Alpha (transparent) component, and R, G, B represents the Red, Green, and Blue components respectively. We also save the corresponding timestamp of the frame for subsequent data synchronization processing. During the measurement, the participants placed their index finger on the rear camera as Figure~\ref{fig:data_collection} illustrates. The finger PPG signal is the spatial average of R-channel values with Max-min normalization from the frames collected by the rear camera. We believe that the contact finger-tip PPG signal is generally very accurate and close to those from the gold standard pulse oximeter.

\subsection{Personalized Algorithm}
\label{sec: personalization_alg}

Camera-based contactless physiological sensing is sensitive to many types of noise, including motion noise, lighting noise (e.g., different ambient lighting), and appearance noise (e.g., different skin types). Due to the complexity of collecting physiological data and potential risks of leaking sensitive information, it is challenging to collect a large-scale camera-based contactless physiological dataset to perform traditional supervised training and train a generalizable model. Moreover, the issues of overfitting in neural network learning based methods have also been raised \cite{chen2018deepphys}. Past research ~\cite{liu2021metaphys} called MetaPhys has demonstrated that meta learning has a great potential in camera-based contactless physiological sensing. MetaPhys has shown that that combing meta learning with a 18s video from the target person can help generate a personalized model for a specific appearance (e.g., skin type). However, this paper did not demonstrate the use of personalization in different lighting conditions and motion tasks and in more ubiquitous and mobile settings. 

In this paper, we adopt and create a variant of MetaPhys to enable few-shot personalization, described in Algorithm 1. Similar to MetaPhys, we also used Model-Agnostic Meta-Learning (MAML) \citep{finn2017model} to update model parameters and create customized models with few-shot learning. The goal of MAML is to produce a robust initialization that enables faster and efficient (i.e., few shots or less data) learning on an unseen target task. Since our algorithm is based on MetaPhys, we apply a similar scheme to train our algorithm. Our algorithm starts with a pre-trained TS-CAN model \cite{liu2020multi} to initialize basic representation to model the spatial and temporal relationship between skin pixels and physiological signals. TS-CAN is a two-branch (appearance and motion) on-device neural network for contactless physiological sensing. The input to TS-CAN is a sequence of video frames (e.g., face) and  the output of it is a sequence of physiological signals ( first-derivative of the pulse signal). The appearance branch takes raw video frames and generates attention masks for the motion branch and helps the motion branch to focus on the regions of interest containing physiological signals instead of unessential parts (e.g., hair, cloth). The motion branch leverages tensor shift module ~\cite{lin2019tsm} to efficiently perform temporal and spatial modeling simultaneously and extract temporal relationships beyond consecutive frames. 

\begin{algorithm}[ht]
\label{alg: mobilephys_alg_train}
\caption{\projectname Training: Meta-learning Algorithm for Mobile Physiological Signal Personalization}
\begin{algorithmic}[1]
\Require $S$: Subject-wise video data
\Require A batch of personalized tasks $\tau$  where each task $\tau_i$ contains N video frames from the front camera and subject $S_i$
\Require A label generator $G$ using rear PPG
\State $\theta \leftarrow$ \textbf{Pre-training} TS-CAN on AFRL dataset
\For {$\tau_i \in \tau $}
\State $K \leftarrow$ Sample $K$ support frames from videos of $\tau_i$ with rear-camera generated contact PPG labels
\State $K\textprime \leftarrow$ Sample $K\textprime$ query frames from videos of $\tau_i$ with rear-camera generated contact PPG labels
\State  $\theta_{\tau_i} \leftarrow \theta - \alpha \nabla_\theta \mathcal L_{\tau_i}f(K, \theta) $, Update the personalized params. based on indiv. support loss
\EndFor
\State $\hat{\theta} \leftarrow \theta - \beta \nabla_{\theta}$ $\sum_{\tau_i} \mathcal L_{\tau_i}f(K\textprime_{\tau_i}, \theta_{\tau_i})$, Update the global params. based on individuals' query loss
\end{algorithmic}
\end{algorithm}

Upon the TS-CAN backbone, we treat each individual as a task and split each individual's data (i.e., facial video data) into a support set $K$ and a query set $K\textprime$. After training with the support set, a personalized model is produced after parameter updates. However, most meta-learning applications assume the labels are available during training, which is not always the case in many machine learning applications. Our novel system, \projectname, can generate self-supervised high-quality "ground truth" PPG signal during the training of meta learning algorithm and produce a personalized model for each unseen individual (task). Therefore, both $K$ and $K\textprime$ come with self-supervised "ground truth" PPG signal labels from finger-tip contact PPG using the rear camera. Since the ultimate goal of meta learning is to generate a robust and efficient initialization, we then use the query set and the self-supervised labels to evaluate the performance of the personalized model  $\theta_i$ and further optimize the global initialization accordingly. The details of the algorithm are described in Algorithm 1.

\section{Data Collection}
In this section, we describe our data collection study, including the apparatus, participant demographic, the user study design and procedure.

\subsection{Apparatus}

\begin{figure*}[ht]
  \centering
  \includegraphics[width=0.85\textwidth]{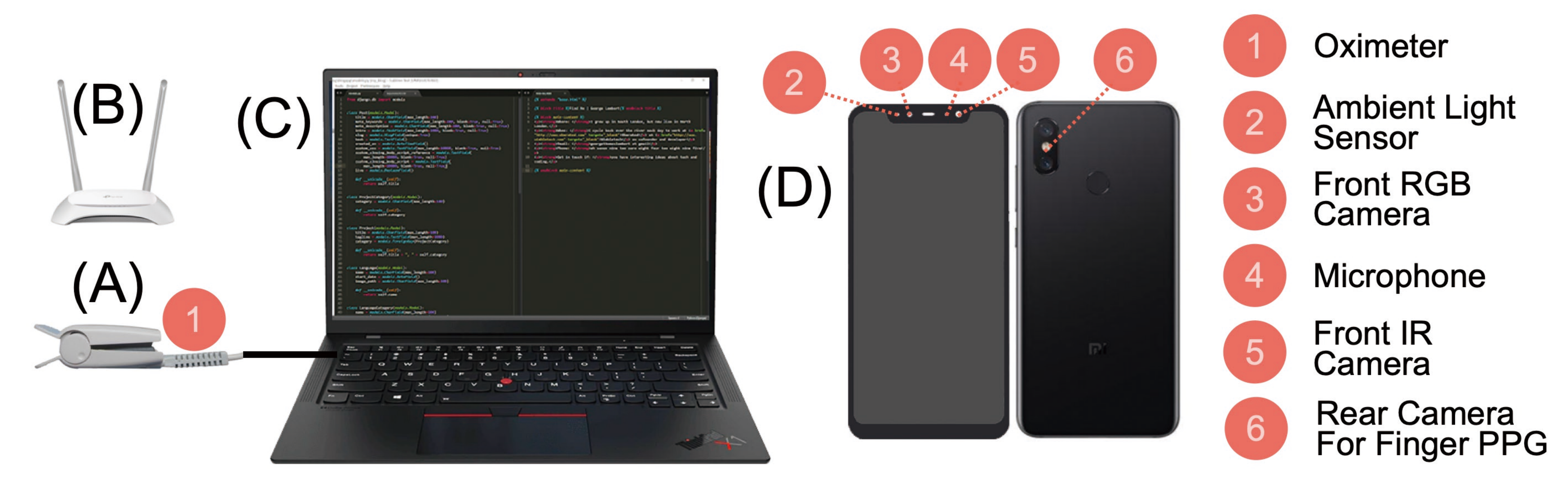}
  \caption{The hardware setup includes (A) an oximeter for gold standard contact PPG measurement, (B) a WiFi router for wirelessly streaming the smartphone's data to the desktop, (C) a desktop with the back-end server for collecting and synchronizing data, and (D) a Xiaomi 8 / iPhone 11 smartphone providing signals from multiple built-in sensors for physiological sensing. }
  \label{fig:apparatus}
\end{figure*}

Figure~\ref{fig:apparatus} shows the apparatus used for data collection. To obtain the gold standard PPG signal, we used a finger pulse oximeter\footnote{HKG-07C+ infrared pulse sensor. http://hfhuake.com/}. The pulse oximeter was connected to a desktop via USB. The raw PPG data were streamed to a desktop running custom python software via the UART protocol. The desktop hosted a back-end server for logging the raw video data from a Xiaomi 8 and an iPhone 11 via a WiFi router. By clicking a start button on a customized data collection application, the mobile app was triggered to simultaneously start the data collection. We recorded raw video from the front RGB camera, the true depth camera (iPhone 11), the ambient light sensor, the microphone, 9-axis inertial measurement unit (IMU), and the rear camera with flash on for camera-based contact gold standard finger PPG. In this work, we only use the RGB videos and pulse oximeter data; however, we anticipate that the other sensor data will be useful in future research and were recorded in our dataset.

\subsection{Participants}
We recruited a total of 39 participants (14 females and 25 males) in age of 20-35 (avg. = 27.1, s.d. = 4.1). Data were collected for 24 subjects using a Xiaomi 8 and for 15 subjects with an iPhone 11. Table~\ref{table:participantsV4} illustrates the distribution of gender and Fitzpatrick skin type~\cite{fitzpatrick1988validity} in our cross-device dataset, as well as the number of subjects who wore glasses, and/or makeup or had facial hair in the video recordings. All the participants are healthy adults and were recruited from a local university. It is worth noting that the second natural light condition is worse in the iPhone 11 data due to lower natural light intensity during the winter months, while the Xiaomi 8 data were collected in the spring/summer.

\begin{figure}[ht]
  \centering
  \includegraphics[width=1\textwidth]{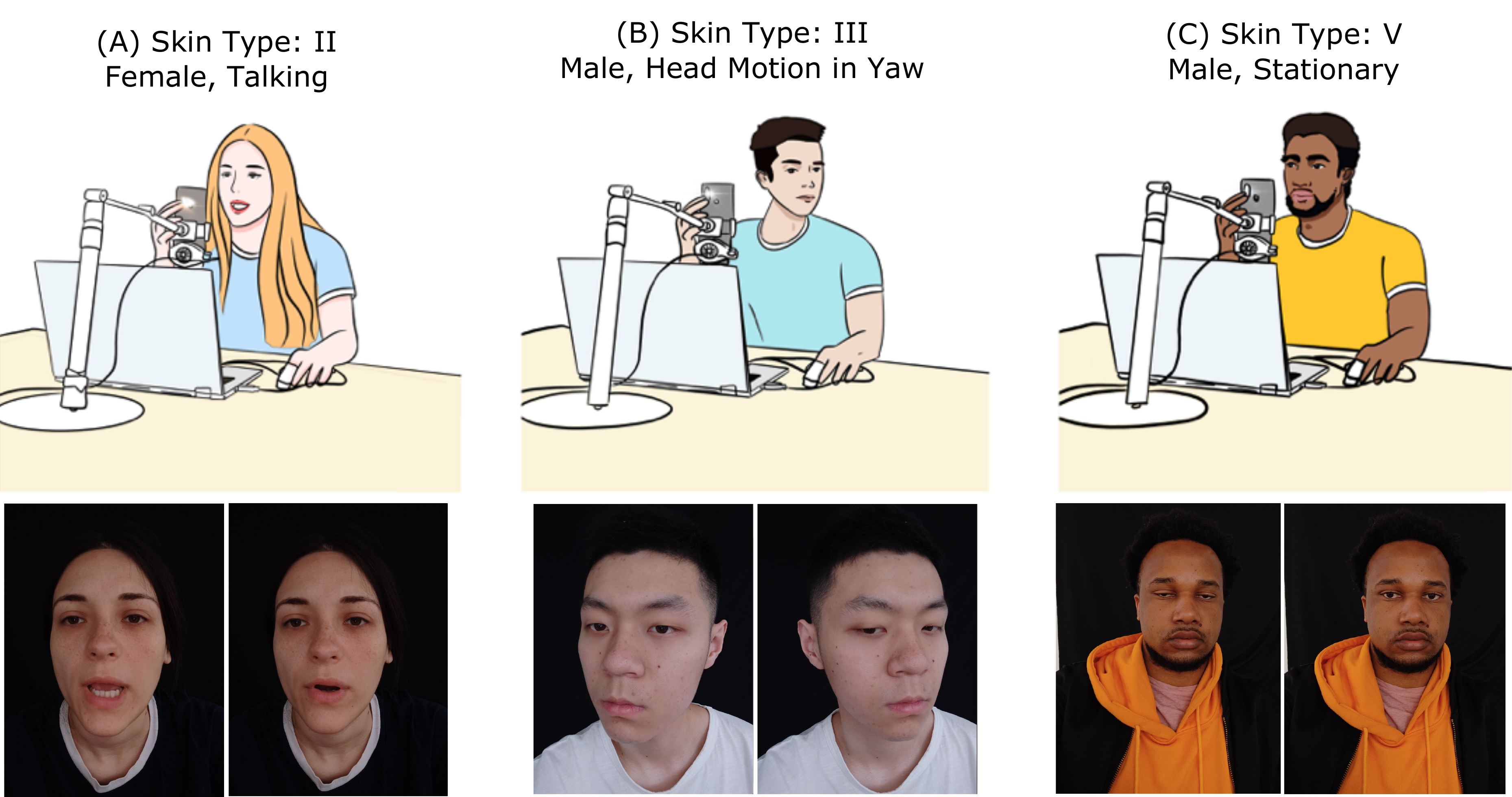}
  \caption{An illustration of some of the tasks in our data collection. We recruited subjects with different skin types and recorded the data under different motion tasks and lighting conditions. The head-shot images show video frames recorded by Xiaomi 8/iPhone 11's front RGB camera.}
  \label{fig:data_collection}
\end{figure}

\begin{table}[ht]
\begin{center}
 \caption{The distribution of gender and Fitzpatrick skin type in our cross-device dataset.}
 \label{tab:overall_subjs}
 \begin{tabular}{c  c  c  c  c  c  c  c  c  c  c  c  c} 
 \toprule
 \textbf{Device} & \textbf{I+II} &  \textbf{III+IV}  & \textbf{V+VI} & \textbf{Female} & \textbf{Male} & \textbf{Glass} & \textbf{Makeup} &  \textbf{Facial Hair} &\textbf{Total}\\
 \toprule
 
 Xiaomi Mi 8 & 8 & 14 & 3 & 7 & 17 & 7 & 5 & 11 & 24 \\
 
 iPhone 11   & 3 & 6 & 5 & 7 & 8 & 4 & 4 & 6 & 15 \\
 
 \hline
 
 All &        11 & 20 & 8 & 14 & 25 & 11 & 9 & 17 & 39 \\
 \bottomrule
 
\end{tabular}
\label{table:participantsV4}
\end{center}
\end{table}

\subsection{Experimental Design}

In this study, we not only explore a personalization approach for adapting contactless PPG measurement models using a dual smartphone camera system \projectname, but also systematically investigate the effect of motion tasks and lighting conditions. Each participant was also recorded before and after exercises. The details of our experiment design are summarized in Table \ref{table:procedureV4} and main variables are discussed in the following:

\begin{itemize}
    \item \textbf{Lighting Condition:} Natural sunlight, LED light, and incandescent light. These three common lights have significant spectrum differences, as Figure 4 shows.
    \item \textbf{Lighting Intensity:} To better investigate how light intensity impacts the performance of camera-based contactless physiological sensing, we also recorded data in three LED lighting intensities (bright(220 Lux)/moderate(110 Lux)/dim(55 Lux)) in the second batch experiment using an iPhone 11. By controlling the luminance value under LED lighting conditions, we were able to mimic the light intensity in different scenarios (220 Lux - office lighting; 110 Lux - family living room lighting; 55 Lux - smartphone screen lighting).
    \item \textbf{Head Motion: }stationary, talking, head rotation in yaw direction, and random head motions as Figure~\ref{fig:data_collection} illustrates.
    \item \textbf{Exercise: } participants were instructed to raise their heart rate by conducting 30 seconds of exercise such as running.
    \item \textbf{Skin Type: } We recruited participants from different backgrounds and have different skin types. The participants are splitted into into three groups based on Fitzpatrick skin type: 1) I+II, 2) III+IV, and 3) V+VI. 
\end{itemize}

\subsection{Experimental Protocol}

The experiments were conducted in a conference room with a large window to ensure the availability of natural ambient light. The natural light could be blocked by closing a thick curtain. 
A desk was placed in the center of the conference room. Participants were asked to sit on the side facing the window. Another black curtain was used as the background screen.  

\begin{figure}[ht]
  \centering
  \includegraphics[width=0.85\textwidth]{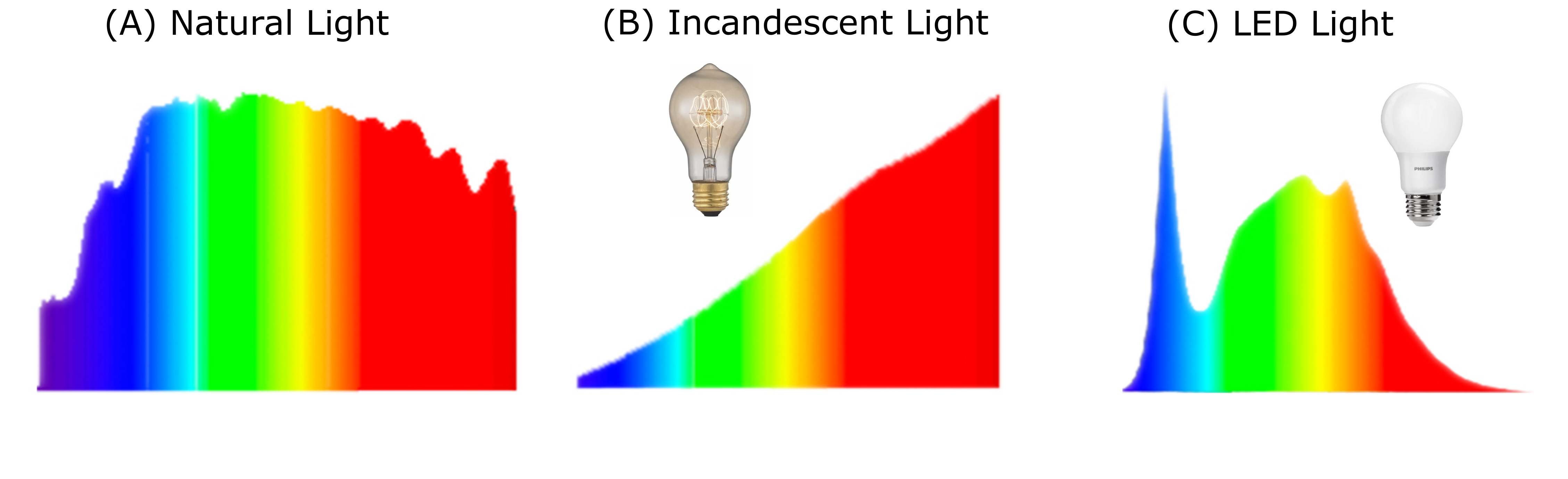}
  \caption{The light spectrum of the three lighting conditions. An incandescent light bulb and a LED lamp are illustrated in (B) and (C). Natural light is broader spectrum than both LED and incandescent illumination.}
  \label{fig:light_spec}
\end{figure}

Every participant was welcomed into the experiment room and informed of the research goals and procedure. After reading and signing the consent form, each participant was instructed to place their left index finger into the pulse oximeter and their right index finger onto the smartphone's rear camera. The position of the smartphone was adjusted to ensure the front camera captured the participant's face. Then, to start each data recording period, the experimenter clicked a button on our customized Android mobile app. This started video recording from the front and rear cameras, the smartphone's sensors (e.g., IMU and microphone) and the contact pulse oximeter. Under the natural sunlight condition, the curtains were adjusted to ensure the facial video was not over dark or exposed. Under LED or Incandescent lighting conditions, curtains were drawn down to 
minimize the sunlight. A LED or incandescent lamp was used to simulate these two lighting scenarios. The distance between the participant and the lamp was carefully adjusted to alter the luminance value measured using a photometer at the participant's face. When participants performed head motions, they were asked to turn their heads at a uniform speed. To explore variable heart rate ranges, we also asked participants to conduct exercises to raise their heart rate on two trials. It took approximately 40 minutes to complete the recordings for each participant. Each participant received a 25 USD gift card.

\begin{table}[ht]
\begin{center}
 \caption{Details of experimental order under different conditions.}
 \label{tab:data_collection}
 \begin{tabular}{c  c  c  c  c  c  c  c  c  c  c} 
 \toprule
 \textbf{Trial No.} & \textbf{Exercise} &  \textbf{Lighting} & \textbf{Motion} & \textbf{Duration (s)}\\ 
 \toprule
 1 & NO & LED (220 Lux - iPhone only) & Stationary & 60\\
 
 2 & NO & LED (110 Lux) & Stationary & 60\\
 
 3 & NO & LED (55 Lux - iPhone only) & Stationary & 60 \\
 
 4 & NO & Incandescent  & Stationary & 60\\

 5 & NO & Natural Sunlight & Stationary & 60\\

 6 & NO & Natural Sunlight & Random & 60\\

 7 & NO & Natural Sunlight  & Yaw Rotation & 60\\

 8 & NO & Natural Sunlight & Talking & 60\\

 9 & YES & Natural Sunlight & Stationary & 60\\

 \bottomrule
 
\end{tabular}
\label{table:procedureV4}
\end{center}
\end{table}

\subsection{Dataset Description}
In total, we collected 168 60-second front camera videos from Xiaomi Mi 8 and 200 front camera videos from iPhone 11. These videos were synchronized with the smartphone's rear-camera based contact PPG signal, true-depth camera signal (iPhone only), 9-axis motion signal (accelerometer, gyroscope, and magnetometer), front ambient light sensor data, and audio signal as well as the oximeter PPG signal. This cross-device dataset was collected to explore multiple methods to enable camera-based contactless physiological sensing using ubiquitous smartphone sensors. In this paper, we only use a subset of the dataset for contactless PPG measurement. Specifically, we utilized facial RGB videos, finger PPG signal and gold standard oximeter PPG signal to explore mobile personalization methods for contactless PPG measurement using the dual-camera setting of the commodity smartphone. However, because of the diversity and high quality of this dataset, researchers will have opportunities to explore other interesting research questions. We plan to release this dataset with this paper.

\section{Training and Evaluation}

\subsection{Training Dataset}

In order to train \projectname, we used two datasets: one for pre-training the backbone (TS-CAN), and the other for training the meta learner $\theta$. We used a similar training regime as MetaPhys~\cite{liu2021metaphys}. We first trained a TS-CAN backbone with the AFRL dataset ~\cite{estepp2014recovering}. This dataset consists of 300 videos from 25 participants, including 17 males and 8 females. The raw resolution of each video is 658x492, and the sampling rate is 30Hz. Gold-standard PPG measurements were recorded from a contact reflective PPG sensor attached to the subject's index finger (similar to the procedure in our study). We use the ground-truth PPG signal from a pulse oximeter to train our backbone. The dataset includes videos with various head motions but not talking. Each participant was asked to keep stationery for the first two trials. Then participants were instructed to perform three head motion tasks, including rotating their head along the vertical axis. The last task of AFRL is random rotation once every second to one of nine predefined locations. For the regular head motion tasks, the subjects were asked to perform different head motions with increasing speed (10 degrees/second, 20 degrees/second, 30 degrees/second, 40 degrees/second, 50 degrees/second, 60 degrees/second).  

Along with the AFRL dataset, we leveraged the UBFC ~\cite{bobbia2019unsupervised} to train the initialization of our meta learner $\theta$. UBFC is a smaller dataset with 42 videos from 42 different participants. The raw resolution of each video is 640x480 in and uncompressed 8-bit RGB format, and the sampling rate is also 30hz. The pulse oximeter they used to collect gold standard PPG signals was a CMS50E transmissive pulse oximeter. Participants were asked to keep stationary during all the experiments. It is worth noting that both AFRL and UBFC are recorded by a professional high-end camera, not a camera from a mobile device such as a smartphone. Neither of these datasets includes subjects with darker skin types (e.g., V, VI). All the data were recorded under one indoor lighting, and the subjects were stationary. 

\subsection{Implementation \& Experimental Details}

\subsubsection{\projectname Implementation: }
We implemented \projectname based on MetaPhys's open-sourced code~\cite{liu2021metaphys}. \projectname is also based on a gradient computation framework called higher \citep{grefenstette2019generalized}. We first trained our backbone network, TS-CAN, using the AFRL dataset. We then train the meta-learner using the UBFC dataset. In this meta-learner training stage, we considered each subject's dataset as a task and used 18 seconds of video data and PPG labels from the pulse oximeter (same as MetaPhys) as the support set $K$ (see \ref{sec: personalization_alg}) for training a personalized model. The rest of the data in the task (i.e., subject) is considered as query data to further evaluate the personalized model and optimize the meta-learner through back-propagation. The output of this training process is a pre-trained meta-learner which only takes 18-second labeled data to generate a personalized model.

In the testing stage, as Algorithm 1 shows, we considered each experimental trial as a task (e.g., natural light + stationary + iPhone 11). The support set is the first 18-second frames recorded from the smartphone's front RGB camera (Xiaomi 8 or iPhone 11) and the ground-truth PPG labels $\tau_i$ were generated by its rear RGB camera. The output of this adaptation process for each task is a personalized model, which can be further used to evaluate the rest of the data within the task. An Adam optimizer \citep{kingma2014adam} and a learning rate of 0.001 were used to optimize the outer loop of Algorithm 1. A stochastic gradient descent (SGD) optimizer with an inner learning rate ($\theta$) of 0.003 was used to optimize the adaption stage. We trained 10 epochs for each experiment.

\subsubsection{Baseline Implementation}
Since prior work has shown that neural network based methods significantly outperform signal processing based demixing approaches. In this work, we only compare \projectname's performance against the state-of-the-art neural method - TS-CAN \cite{liu2020multi} and MetaPhys \cite{liu2021metaphys}. Our goal is to propose a mobile physiological sensing system; therefore, using models that run on-device is important. To the best of our knowledge, TS-CAN and MetaPhys are the best baseline papers that focus on on-device camera-based physiological sensing. Therefore, we chose TS-CAN and MetaPhys as our baselines.

For TS-CAN experiments, we trained TS-CAN with aggregated training datasets of AFRL and UBFC as the backbone network. In the testing stage of each task (same experiment trial as the \projectname's testing stage), we then use the first 18 second's video data and label to fine tune the network and evaluate the rest of the data within the task.

For MetaPhys experiments, we used the same UBFC pretrained meta-learner described in the previous section. However, during the teasing stage, the label used for adaption in MetaPhys is pseudo labels generated by POS \citep{wang2016algorithmic} according to MetaPhys. POS is an unsupervised signal processing method that computes a projection plane orthogonal to the skin type based on optical and physiological principles. The difference between \projectname and MetaPhys is the label used for few-shot adaption where our proposed \projectname is able to generate high-quality pseudo labels while MetaPhys relies on the unreliable pseudo label, which could be very noisy in challenging tasks (e.g., darker skin types, lighting, motion, etc.)

After getting the predicted PPG waveform from the network, we applied a band-pass filter to remove unessential noise. More superficially, we used a 2nd-order butterworth filter with a cutoff frequency of 0.75 and 2.5 Hz to keep the signal containing realistic adults' heart rate. During the evaluation stage, we compare the heart rate computed from filtered predicted PPG signal against the heart rate calculated from PPG labels recorded by a pulse oximeter (see Figure \ref{fig:mobilephys}). 

\subsection{Evaluation Metric}

We used three metrics to evaluate the performance of \projectname and our baseline. 

\subsubsection{Mean Absolute Error (MAE)}

For calculating the MAE between our model estimates and the gold-standard heart rate calculated from the contact PPG sensor across all the subjects within our collected dataset. The equation of MAE is shown below:

\begin{equation}
MAE = \frac{1}{T}\sum_{i=1}^{T} |HR_{i} - HR'_{i}|
\end{equation}

In Equation 1, HR is the gold-standard heart rate and HR' is the estimated heart rate from the videos, respectively. The gold-standard HR frequency was calculated from the PPG signal collected from a pulse oximeter. 

\subsubsection{Pulse Signal-to-Noise Ratios (SNR)}

We calculated PPG signal-to-noise ratios (SNR) based on the method proposed by De Haan et al.~\citep{de2013robust}. This captures the signal quality of the recovered predicted camera-based contactless PPG signal. The gold-standard HR frequency was determined from the contact pulse oximeter. 

\begin{equation}
SNR = 10\mathrm{log}_{10}\left(\frac{\sum^{240}_{f=30}((U_t(f)\hat{S}(f))^2}{\sum^{240}_{f=30}(1 - U_t(f))\hat{S}(f))^2)}\right)
\end{equation}

where $\hat{S}$ is the power spectrum of the PPG signal (S), $\textit{f}$ is the frequency in beats per minute. HR is the heart rate calculated from the gold-standard device and $\textit{U$_{t}$(f)}$ is a binary template for the heart rate region from HR-6 BPM to HR+6BPM and its first harmonic region from 2*HR-12BPM to 2*HR+12BPM, and 0 elsewhere. These parameters and regions are defined by in \citep{de2013robust}. 

\subsubsection{Pearson Coefficient}

We also calculated the Pearson Correlation between the estimated heart rates from camera-based contactless physiological measurement and the gold-standard heart rates from the contact pulse oximeter.

\section{Results and Findings}
\label{sec:result}
\subsection{Quantitative Results of \projectname}
\begin{table*}[t]
    \centering
	\footnotesize
	\caption{Pulse Measurement (Heart Rate) on Different Motion Tasks.}
	\label{tab:result_motion_task}
	\setlength\tabcolsep{3pt} 
	\begin{tabular}{c|ccc|ccc}
	\toprule
		 & \multicolumn{3}{c}{\textbf{Experimental Conditions}} & \multicolumn{3}{c}{\textbf{Xiaomi 8 / iPhone 11}}\\
        \multicolumn{1}{c}{Method} & Exercise & Task & Lighting & MAE & SNR & $\rho$  \\ \hline \hline
        \textbf{\projectname} & No & Head Motion in Yaw & Natural Light & \textbf{10.71}/\textbf{11.40} & \textbf{-8.73}/\textbf{-8.05} & \textbf{0.40}/\textbf{0.34}  \\
        TS-CAN & No & Head Motion in Yaw & Natural Light & 12.04/12.89 & -9.91/-9.46 & 0.06/0.33  \\
        MetaPhys & No & Head Motion in Yaw & Natural Light & 13.35/13.93 & -8.81/-12.77 & 0.37/0.21  \\
        \hline
        \textbf{\projectname} & No & Talking & Natural Light & \textbf{3.38}/\textbf{6.33} & \textbf{-2.16}/\textbf{-7.61} & \textbf{0.70}/\textbf{0.59}  \\
        TS-CAN & No & Talking & Natural Light & 11.61/9.38 & -7.05/-10.53 & 0.04/0.33  \\
        MetaPhys & No & Talking & Natural Light & 3.99/7.67 & -3.24/-9.93 & 0.65/0.43  \\
        \hline
        \textbf{\projectname} & No & Random Head Motion & Natural Light & \textbf{4.74}/\textbf{6.27} & \textbf{-6.47}/\textbf{-7.04} & \textbf{0.84}/\textbf{0.58}  \\
        TS-CAN & No & Random Head Motion & Natural Light & 13.44/7.14 & -11.9/-7.68 & 0.31/\textbf{0.58}  \\
        MetaPhys & No & Random Head Motion & Natural Light & 8.92/6.98 & -8.69/-8.25 & 0.33/0.56  \\
        \hline
        \bottomrule 
  \end{tabular}
    \\
  \tiny{MAE = Mean Absolute Error, $\rho$ = Pearson Correlation, SNR = BVP Signal-to-Noise Ratio.}
  \vspace{-0.4cm}
\end{table*}

\begin{table*}[t]
    \centering
	\footnotesize
	\caption{Pulse Measurement (Heart Rate) after Exercising.}
	\label{tab:exercise_table}
	\setlength\tabcolsep{3pt} 
	\begin{tabular}{c|ccc|ccc}
	\toprule
		    & \multicolumn{3}{c}{\textbf{Experimental Conditions}} & \multicolumn{3}{c}{\textbf{Xiaomi 8 / iPhone 11}}\\
        \multicolumn{1}{c}{Method} & Exercise & Task & Lighting & MAE & SNR & $\rho$  \\ \hline
        \hline
        \textbf{\projectname} & Yes & Stationary After Exercise & Natural Light & \textbf{3.49}/\textbf{7.68} & -2.67/\textbf{-7.43} & \textbf{0.85}/\textbf{0.79}  \\
        TS-CAN & Yes & Stationary After Exercise & Natural Light & 8.67/10.83 & -5.36/-9.75 & 0.44/0.53  \\
        MetaPhys & Yes & Stationary After Exercise & Natural Light & 3.69/12.63 & \textbf{-2.06}/-10.79 & 0.81/0.33  \\
        \hline
        \bottomrule 
  \end{tabular}
    \\
  \tiny{MAE = Mean Absolute Error, $\rho$ = Pearson Correlation, SNR = BVP Signal-to-Noise Ratio.}
\end{table*}

\begin{table*}[t!]
    \centering
	\footnotesize
	\caption{Pulse Measurement (Heart Rate) on Different Lighting Conditions.}
	\label{tab:lighting_table}
	\setlength\tabcolsep{3pt} 
	\begin{tabular}{c|ccc|ccc}
	\toprule
	    & \multicolumn{3}{c}{\textbf{Experimental Conditions}} & \multicolumn{3}{c}{\textbf{Xiaomi 8 / iPhone 11}}\\
        \multicolumn{1}{c}{Method} & Exercise & Task & Lighting & MAE & SNR & $\rho$  \\ \hline \hline
        \textbf{\projectname} & No & Stationary & Natural Light & \textbf{0.97}/\textbf{4.27} & \textbf{1.54}/\textbf{-7.04} & \textbf{0.99}/\textbf{0.71}  \\
        TS-CAN & No & Stationary & Natural Light & 4.32/7.14 & -0.31/-7.68 & 0.49/0.58\\
        MetaPhys & No & Stationary & Natural Light & 1.49/6.98 & 0.24/-8.25 & 0.97/0.56 \\
        \hline
        \textbf{\projectname} & No & Stationary & Incandescent & \textbf{0.73}/\textbf{5.50} & 2.87/\textbf{-7.33} & \textbf{0.99}/\textbf{0.85}   \\
        TS-CAN  & No & Stationary & Incandescent & 1.05/10.72 & 1.63/-10.76 & 0.97/0.02 \\
        MetaPhys  & No & Stationary & Incandescent & 2.60/10.34 & \textbf{3.06}/-9.91 & 0.73/0.43 \\
        \hline
        \textbf{\projectname} & No & Stationary & LED & 1.76/\textbf{3.05} & \textbf{2.85}/\textbf{-1.99} & 0.96/\textbf{0.86}  \\
        TS-CAN & No & Stationary & LED & \textbf{1.63}/8.37 & 2.22/-5.07 & \textbf{0.99}/0.09  \\
        MetaPhys & No & Stationary & LED & 1.65/7.69 & 2.56/-7.58 & 0.96/0.54  \\
 
        \hline
        \bottomrule 
  \end{tabular}
    \\
  \tiny{MAE = Mean Absolute Error, $\rho$ = Pearson Correlation, SNR = BVP Signal-to-Noise Ratio.}
\end{table*}

\begin{table*}[t!]
    \centering
	\footnotesize
	\caption{Pulse Measurement (Heart Rate) on Different Skin Types. All the participants are used to evaluate how skin types impact \projectname. }
	\label{tab:skin_tone_table}
	\setlength\tabcolsep{3pt} 
	\begin{tabular}{c|ccc|ccc}
	\toprule
	    & \multicolumn{3}{c}{\textbf{Experimental Conditions}} & \multicolumn{3}{c}{\textbf{All Subjects}}\\
        \multicolumn{1}{c}{Method} & Skin & Task & Lighting & MAE & SNR & $\rho$  \\ \hline \hline
        \textbf{\projectname} & I+II & Stationary & Natural Light & \textbf{1.51} & \textbf{0.16} & \textbf{0.98}  \\
        TS-CAN & I+II & Stationary & Natural Light & 4.52 & -1.92 & 0.60\\
        MetaPhys & I+II & Stationary & Natural Light & 2.19 & -0.95 & 0.95 \\
        \hline
        \textbf{\projectname} & III+IV & Stationary & Natural Light & \textbf{1.04} &\textbf{0.09} & \textbf{0.99}   \\
        TS-CAN  & III+IV & Stationary & Natural Light & 2.69 & -1.18 & 0.91 \\
        MetaPhys  & III+IV & Stationary & Natural Light & 1.91 & -1.04 & 0.88 \\
        \hline
        \textbf{\projectname} & V+VI & Stationary & Natural Light & \textbf{2.31} & \textbf{-7.30} & \textbf{0.98}  \\
        TS-CAN & V+VI & Stationary & Natural Light & 8.45 & -8.63 & 0.68  \\
        MetaPhys & V+VI & Stationary & Natural Light & 17.31 & -8.65 & 0.42  \\
 
        \hline
        \bottomrule 
  \end{tabular}
    \\
  \tiny{MAE = Mean Absolute Error, $\rho$ = Pearson Correlation, SNR = BVP Signal-to-Noise Ratio.}
\end{table*}

\begin{table*}[t!]
    \centering
	\footnotesize
	\caption{Pulse Measurement (Heart Rate) on Different LED Light Intensities. Only participants with iPhone 11 were enrolled in light intensity studies.}
	\label{tab:lighting_level_table}
	\setlength\tabcolsep{3pt} 
	\begin{tabular}{c|ccc|ccc}
	\toprule
	    & \multicolumn{3}{c}{\textbf{Experimental Conditions}} & \multicolumn{3}{c}{\textbf{iPhone Subjects}}\\
        \multicolumn{1}{c}{Method} & Light Intensity & Task & Lighting & MAE & SNR & $\rho$  \\ \hline \hline
        \textbf{\projectname} & 55 Lux & Stationary & LED & \textbf{4.58} & \textbf{-6.24} & \textbf{0.87}  \\
        TS-CAN & 55 Lux & Stationary & LED & 6.78 & -7.91 & 0.14\\
        MetaPhys & 55 Lux & Stationary & LED & 6.98 & -9.40 & 0.80 \\
        \hline
        \textbf{\projectname} & 110 Lux & Stationary & LED & \textbf{3.27} & -3.03 & \textbf{0.81}   \\
        TS-CAN  & 110 Lux & Stationary & LED & 6.28 & -4.79 & 0.36 \\
        MetaPhys  & 110 Lux & Stationary & LED & 6.26 & \textbf{-2.55} & 0.12 \\
        \hline
        \textbf{\projectname} & 220 Lux & Stationary & LED & 1.93 & -1.59 & 0.97  \\
        TS-CAN & 220 Lux & Stationary & LED & \textbf{1.64} & \textbf{-1.38} & \textbf{0.98}  \\
        MetaPhys & 220 Lux & Stationary & LED & 4.94 & -5.59 & 0.65  \\
        \hline
        \bottomrule 
  \end{tabular}
    \\
  \tiny{MAE = Mean Absolute Error, $\rho$ = Pearson Correlation, SNR = BVP Signal-to-Noise Ratio.}
\end{table*}

We compare the performance of \projectname with the state-of-the-art traditional supervised training algorithm (TS-CAN) and model-agnostic meta-learning algorithm (MetaPhys) using the data collected from Xiaomi 8 (Android) and iPhone 11 (iOS). The reported measurements include Mean-Square-Error (MAE), Signal-to-Noise Ratio (SNR) and Pearson Coefficient as metrics ($\rho$). In order to investigate how different motion tasks impact the performance of camera-based contactless physiological sensing systems, we first conducted experiments when the lighting condition was fixed. As Table \ref{tab:result_motion_task} shows, under the natural light condition, we compared the performance of \projectname and baselines in three motion tasks: 1) head motion in yaw, 2) talking and 3) random head motion. On the head of motion in yaw task, \projectname improved 11.3\%/19.1\% in MAE, 13.0\%/18.0\% in SNR, and 130.0\%/22.2\% in Pearson coefficient in average across all the subjects in two mobile devices when compared with TS-CAN and MetaPhys respectively. On the talking task, \projectname outperformed our baselines by 58.0\%/16.5\% in MAE, 49.3\%/26.8\% in SNR, and 334.0\%/16.3\% in Pearson coefficient, compared to our baseline methods. For the random head motion task, \projectname showed it could enhance the performance by 51.6\%/34.8\% in MAE, 34.9\%/21.5\% in SNR and 78.8\%/76.8\% in Pearson coefficient.

Furthermore, to help us understand whether the models work well for higher heart rates, we also evaluated \projectname on the recordings collected immediately after the participants conducted one minute exercise (e.g., running). As Table \ref{tab:exercise_table} shows, \projectname demonstrates superior performance compared to TS-CAN and MetaPhys where it can reduce the MAE by 46.3\%/28.4\%. The SNR and Pearson Coefficient are also improved by 36.1\%/20.0\% and 74.2\%/32.2\%, respectively.

Next, we explored the effects of lighting conditions that potentially could have a large impact on the performance of mobile camera-based contactless physiological sensing systems. We conducted three sets of experiments to examine three different lighting conditions: 1) natural sunlight, 2) incandescent light, and 3) LED light. All the experiments were conducted on the video recordings collected while the participants were stationary to exclude confounds from motion. As Table \ref{tab:lighting_table} illustrates, in the natural light condition, \projectname improves 58.6\%/37.8\% in MAE, 44.0\%/41.8\% in SNR and 68.2\%/8.6\% in pearson coefficient. In the incandescent light condition, \projectname enhances 46.2\%/54.0\% in MAE, 66.4\%/45.4\% in SNR and 54.8\%/52.3\% in pearson coefficient. In the LED light condition, \projectname enhances 46.6\%/43.2\% in MAE, 269.3\%/173.8\% in SNR and 43.1\%/15.4\% in pearson coefficient. Besides of different types of lighting conditions, we also evaluated \projectname on different lighting intensities on LED light (e.g., dimmer light). As Table \ref{tab:lighting_level_table} illustrates, in the 55 Lux condition, \projectname outperforms the baseline methods by 32.4\%/31.5\% in MAE, 21.5\%/33.7\% in SNR and 520\%/9\% in the pearson coefficient regarding the light intensity. In the 110 lux setting, we also observed a similar trend where \projectname outperforms both baseline methods in MAE and Pearson Coefficient. However, in the Lux 220 settings, \projectname achieves similar performance as TS-CAN while MetaPhys failed to generalize well in this setting.

To ensure our proposed system does not have a bias on a specific population or race, we evaluated \projectname's effectiveness across different skin types as Table~\ref{tab:skin_tone_table} shows. In this set of experiments, we combined the subjects from Xiaomi 8 and iPhone 11 to get more balanced skin distribution across skin types. Results show that \projectname outperforms the baseline methods by 66.6\%/31.1\% in MAE, 108\%/117\% in SNR and 63.3\%/3.1\% in the pearson coefficient regarding the skin type I and II; 61.3\%/45.5\% in MAE, 108\%/109\% in SNR and 8.8\%/12.5\% in the pearson coefficient regarding the skin type III and IV; and 72.7\%/86.7\% in MAE, 15.4\%/15.6\% in SNR and 44.1\%/133\% in the pearson coefficient regarding the skin type V and VI.

\subsection{Findings of Mobile Camera Settings}

We observed that the Android phone camera settings greatly affected the quality of the video as well as the overall performance of camera-based physiological sensing. Anecdotally, we observed similar behavior in iPhone. Therefore, we conducted a systematic analysis on various camera settings on Xiaomi Mi 8 to explore how mobile camera settings affect the performance of contactless PPG measurement. 

We choose the three most common camera settings that are: auto-white-balance (AWB), exposure time and sensitivity. Since camera-based physiological sensing systems aim to extract very subtle color changes on the skin, those settings play a significant role in the RGB values captured by the camera. More explicitly, these settings together determine the brightness and color of the video and the intake of the light from the camera lens. To further complicate matters, in video recordings, these parameters are typically changed dynamically and are not constant for a single video.

In the Android API, two parameters: color correction gains and color correction transform determine the AWB algorithm. Color correction gains are gains applying to Bayer raw color channels for white-balance, and color transform is a matrix used to transform the sensor RGB color space to the output linear sRGB color space. These two values are automatically generated when AWB is on, while in manual mode, users set these two parameters manually.

We conducted 13 experiments to examine the influence of auto-white-balance, camera sensitivity and exposure time on the performance of our camera-based physiological sensing system. The experiment procedure and the results are shown in Table \ref{table:camerasetting}. We repeated each experiment five times using the same camera setting, and the average heart rate MAEs are reported. All the experiments were completed in an hour to ensure that the lighting conditions were consistent across different experiments. In experiment \#1, when auto exposure and auto white balance are turned on, a sensitivity of 175  exposure time of 1/30s was obtained as typical values and was used as the basis of the subsequent experiments. Typical color correction gains and color correction transform values were also obtained in this experiment and used for manual AWB mode. In the experiments \#2-\#10, auto exposure was turned off. Moreover, we also explored the sensitivity-exposure time space across nine different experiments. Experiment \#11-\#13 adopted the same exposure time and sensitivity as experiment \#7, \#5 and \#8. The AWB parameters were set manually to see if AWB has a positive impact on the video quality,

In Table \ref{table:camerasetting}, the results show a noticeable decrease in averaged MAE when exposure time and sensitivity are set to a lower value. This indicates that these settings controlled the light admitted by the system and made the videos relatively darker. Camera-based contactless physiological measurement aims to extract subtle changes ($\delta$) of skin pixels; therefore, making the video too bright could corrupt these subtle signals. However, the averaged MAE in experiment \#10 was increased sharply when we continued lowering the exposure time. This increment of MAE indicates that forcing the video to be too dark also makes extraction of physiological signals challenging as some of the individual's facial details were not even visible. Therefore, we set the exposure time to 1/50 s and sensitivity to 100 to balance the brightness of videos. As for AWB, in all three experiments where AWB is off, the average MAE is slightly higher, which indicates that AWB helps capture higher quality of videos for the camera-based contactless physiological system. These findings in mobile camera settings help our system to get the best quality of videos recorded by smartphones' cameras. 

Moreover, we also observed similar results on an OPPO Reno 5 smartphone, which suggests that this pattern is not device-dependent. This may guide future studies to set the proper mobile camera settings for camera-based contactless physiological sensing.

\begin{table}[h!]
\begin{center}
\caption{Experiments on exploring the effect of smartphone front camera settings on the pulse measurement performance.}
\begin{adjustbox}{width=\columnwidth,center}
\label{table:camerasetting}
 \begin{tabular}{c c c  c  c  c  c  c  c} 
 \toprule
 \textbf{Trial No.} & \textbf{Duration (s)} & \textbf{Auto White Balance} & \textbf{Auto Exposure} &  \textbf{Exposure time (s)} & \textbf{Sensitivity} & \textbf{MAE (beats/min)}\\ 
 \toprule
     1 & 30 & On & On & Auto & Auto & 5.6 \\
     
     2 & 30 & On & Off & 1/30 & 175 & 2.9\\
     
     3 & 30 & On & Off & 1/30 & 100 & 1.2\\
     
     4 & 30 & On & Off & 1/30 & 250 & 5.4\\
     
     5 & 30 & On & Off & 1/50 & 175 & 2.5\\
     
    6 & 30 & On & Off & 1/100 & 175 & 0.8\\
    
    7 & 30 & On & Off & 1/50 & 250 & 2.5\\
    
    8 & 30 & On & Off & 1/50 & 100 & 0.5\\
    
    9 & 30 & On & Off & 1/100 & 250 & 0.6\\
    
    10 & 30 & On & Off & 1/100 & 100 & 3.9\\
    
    11 & 30 & Off & Off & 1/50 & 250 & 3.1\\
    
    12 & 30 & Off & Off & 1/50 & 175 & 2.8\\
    
    13 & 30 & Off & Off & 1/50 & 100 & 1.3\\
 \bottomrule
 
\end{tabular}
\end{adjustbox}
\end{center}
\end{table}

\section{Discussion}
In this paper, we demonstrated the feasibility of our proposed mobile personalizing camera-based contactless physiological sensing system using the power of meta learning, without the need for a clinical-grade sensor. This is achieved by using both the front and rear cameras on a smartphone to generate high-quality synchronized self-supervised labels. These labels are then used for training personalized contactless camera-based PPG models. These personalized models could then be shared with other devices. We foresee the opportunity of future IoT applications (e.g., smart mirrors or other fitness applications) utilizing the smartphone's rear camera for a short calibration of a person's appearance and environment to personalize the physiological model. In the following sections, we discuss our major findings, limitation, and future work. 

\subsection{How Does \projectname Compare with the Baseline Methods?}

Based on the results shown in Table \ref{tab:result_motion_task}-\ref{tab:lighting_level_table}\, \projectname consistently outperforms the baseline methods across the tasks, with very few exceptions. Specifically, it helps significantly on the motion tasks (e.g., talking and random head motion) and in measurement across skin types based on Table \ref{tab:result_motion_task} and Table \ref{tab:skin_tone_table}. We find that \projectname achieves similar performance to MetaPhys in the talking and stationary after exercising tasks, but MetaPhys performs significantly worse than \projectname and TS-CAN in the random head motion, incandescent light and skin type (V+VI) tasks. We believe this is because MetaPhys is highly reliant on the POS method \cite{wang2016algorithmic} to generate high-quality pseudo labels, but POS, a signal separation method fails to yield accurate PPG waveform in some of these more challenging tasks.

\subsection{How Does \projectname Perform in iOS and Android?}
Based on Table \ref{tab:result_motion_task}-\ref{tab:lighting_level_table}, \projectname sees similar performance improvements on both Xiaomi 8 (Android) and iPhone 11 (iOS). These results indicate that \projectname has the potential to generalize to different mobile devices. Overall, we see no reason that the method would not provide benefits on other mobile devices, given that imaging devices on almost all mid-range and above smart phones have good specifications. However, while there are improvements over the baselines, we do observe that errors for subjects using the iPhone 11 are higher than those for subjects using Xiaomi 8. After analyzing the results, we found that there were two reasons causing this difference: 1) the data from iPhone 11 were collected over the winter quarter; therefore, there was a lower intensity of natural light in the recording environments than for the Xiaomi 8 data, which was collected over the spring/summer. It is harder to capture subtle skin pixel changes in low lighting settings. 2) The iPhone 11 data include 5 subjects with skin types V+VI (1 V + 4 VI). Those subjects have higher melanin content in their skin, which impacts the intensity of light reflected from the body and the amount of light captured by a camera. Therefore, extracting subtle pixel changes from darker skin type subjects is more challenging~\cite{nowara2020meta}. In general, the iPhone 11 batch is simply a more challenging dataset than the Xiaomi 8 batch. We do not think that these differences are as much due to the hardware as these other factors. 

\subsection{How Do Motion Tasks Impact Camera-based Contactless Physiological Systems?}

As Table \ref{tab:result_motion_task} shows, \projectname leads to considerable performance gains on tasks with larger motions  (taking and random head motion). However, despite \projectname achieving more than 10\% improvement in MAE of the task of head motion in yaw, it is failed to remove the large noise introduced by the larger motions as the MAE for this task even after personalization is still quite large. The subjects had larger motion in this task compared to the talking and random head motion tasks. These results indicate that it is easier for the model to pick up the personalized features in relatively stationary video frames. Large motions bring extra noise and \projectname may get ``confused'' about what to learn during the personalization phase. Moreover, UBFC only contains videos of stationary subjects; therefore, our meta-learner was only trained to learn how to adapt to relatively stationary tasks. The magnitude of motion was smaller in the tasks of talking and random head motion, than in yaw, and \projectname is able to yield greater improvements on these tasks.

\subsection{How Do Lighting Conditions Impact Camera-based Contactless Physiological Systems?}

Based on the results in Table \ref{tab:lighting_table}, \projectname yields superior performance on the conditions of natural light and incandescent light. \projectname reduces errors in the videos with incandescent light by approximately 40\% on average across all the subjects. It is not surprising that \projectname achieves such results because none of the videos in AFRL and UBFC are recorded under incandescent lighting and incandescent has a different spectral composition compared to other lighting (e.g., sunlight), as illustrated in Figure \ref{fig:light_spec}. MetaPhys also does not work well in such complex lighting conditions, as POS was not designed to handle videos with complex lighting environments. Furthermore, we observe that \projectname provides even more benefit for the videos under natural light.  Natural sunlight has a broad spectrum (see Figure \ref{fig:light_spec}-A), and we hypothesize that our training data failed to represent a complete spectrum of natural sunlight. Therefore, performing a few-shot personalization substantially reduces the error by helping the model adapt to the different spectral profiles.

On the other hand, \projectname, MetaPhys and TS-CAN achieved similar performance under LED lighting. Since our training data (AFRL and UBFC) have a similar lighting spectral profile as the videos we collected under LED lighting, TS-CAN already performs strongly, showing that there is no need to personalize the model if the training data includes similar lighting conditions. However, in practice it is unlikely that the exact lighting conditions will be known at training time.

\subsection{How Does Exercise Impact Camera-based Contactless Physiological Systems?}

It is clear that \projectname achieves better performance than TS-CAN after subjects raise their heart rate by exercising. Heart rate can be dramatically elevated after exercise. However, most of our training data have a regular range of heart rate (60 BMP to 80 BMP). MetaPhys also achieves similar performance as \projectname because the task was relatively stationary and under good lighting conditions so that POS was able to generate good pseudo PPG labels. Our results suggest that personalization is important when camera-based contactless physiological sensing technologies are deployed in fitness settings.

\subsection{How Does Skin-type Impact Camera-based Contactless Physiological Systems?}

As Table \ref{tab:skin_tone_table} shows,  \projectname achieves superior performance, especially in the subjects who have darker skin types such as VI. It is not surprising that MetaPhys also helps to provide an improvement on subjects of I, II,  III and IV because of the personalization process. However, when it comes to the subjects in categories V+VI, MetaPhys failed to generalize to these subjects because POS cannot yield reliable pseudo PPG waveforms for the meta-learner. Previous work has already highlighted this issue with POS~\cite{nowara2020meta}. MetaPhys was even beaten by TS-CAN, and we believe it is because the meta-learner used an inaccurate waveform to generate the personalization model. On the other hand, \projectname was still able to provide nearly 75\% improvement compared to pre-trained TS-CAN. Through these results, we believe \projectname will be specifically useful in improving the equability of camera-based contactless physiological sensing.

\subsection{Limitations}

Although \projectname demonstrates it is able to achieve superior results in complex conditions, there remain limitations: (1) During data collection the smartphone was fixed on a stand. This helped us reduce some subtle motion artifacts. It might be reasonable to ask users to place/hold the phone for 18 seconds during the deployment; however, it would not be trivial to guarantee that they do so. (2) Although \projectname can help reduce the performance gaps between people with different skin types, there still remains some disparities. As Table \ref{tab:result_motion_task}-\ref{tab:lighting_level_table} show, \projectname achieved higher errors on the iPhone 11 dataset than Xiaomi 8 dataset. We hypothesize this is because there are more VI subjects in the iPhone dataset. Our proposed method cannot completely close the gap, although we believe that our approach is a step in the right direction; (3) We are aware that we have a non-uniform distribution of skin types in this dataset, and the same is true for many other PPG datasets~\cite{nowara2020meta}. Specifically, we only have eight participants with the skin type of VI and V. Recent efforts have been made to address these imbalances, but these data are not publicly available at this time~\cite{mcduff2020advancing,chari2020diverse}. We plan to expand our dataset with better coverage of skin types. (4) Finally, our system was only evaluated on limited daily motion tasks such as talking, and it is worth collecting more data with more routine activities such as typing, walking, etc.  However, recruitment was challenging during the COVID-19 pandemic and we tried to capture a range of environments, activities and demographics; (5) The current system requires running a few-shot personalization process on every single task, which means users need to calibrate the system when they change their environment or activity.

\subsection{Future Work}

We are pushing toward more robust and generalizable mobile physiological sensing by demonstrating \projectname's performance on videos with two mobile devices, large head motions, different ambient lighting conditions and mobile camera settings. However, we conducted our experiments in a lab environment, and simulating different lighting conditions does not reflect the full diversity of conditions observed in everyday life. To further enable practical mobile physiological sensing, we would expect future work to study other mobile settings, such  as deploying camera-based contactless physiological sensing in outdoor environments or in a gym. Moreover, modern smartphones are equipped with advanced cameras such as true-depth, IR cameras. It is also worth exploring how to leverage these advanced sensors to perform physiological sensing. 

We noticed a 1-Hz noise signal that was constantly an issue when we collected the facial videos using the front RGB camera. We have noticed that this issue is not isolated only to the Xiaomi Mi 8 smartphone model. We observed the same issues on three other brands of Android smartphones and on iPhone with different camera specifications. Although we were not fully aware of the reason behind the 1-Hz noise issue, we observed that it was more likely to occur when the smartphone started to heat up and when using a large ISO setting and a longer exposure. Therefore, to ensure the quality of the facial videos, it is better to set a smaller ISO, use a faster shutter. We adopted a cooler and switched smartphones between recordings to ensure the phone did not overheat. Finally, we double-checked all recordings at the end of each experiment. We re-collected the data if we observed a significant 1-Hz noise issue in specific recordings. This issue seems to be partially related to the hardware and not entirely to our processing of the video.

We collected a large multi-modality mobile physiological sensing PPG dataset, which will be released with this paper. We would expect future work to explore novel contactless or contact physiological sensing methods and applications using our dataset. We foresee the opportunities of a multi-modality sensing approach (e.g., IMU, Audio, Ambient light and RGB videos, etc.), contactless PPG measurement using the front IR cameras, computing other physiological signals (e.g., respiratory rate or heart rate variability), and other physiological computing applications.

\section{Conclusion}

Camera-based contactless physiological sensing holds promise for monitoring vital signs and important cardiac and pulmonary parameters. For example, these systems could provide more comfortable and convenient ways to screen for arrhythmias and diagnose and monitor atrial fibrillation as well as cardiopulmonary diseases. In this paper, we present a novel mobile camera-based contactless physiological sensing system called \projectname, that leverages front and back cameras to provide self-supervised ``ground truth'' labels to our few-shot meta learning algorithm to perform personalization and environmental adaptation. To validate the robustness of our system, we also release the first-ever multi-modality mobile remote physiological dataset with different mobile devices, lighting conditions, motions, activities, and skin types. Our superior results demonstrate our proposed system substantially improves over the state-of-art system under different contexts. Furthermore, we systematically examined how camera settings in smartphones impact the performance of camera-based contactless physiological sensing, which will provide useful guidance for other researchers who are building smartphone-based contactless physiological sensing systems.

\bibliographystyle{ACM-Reference-Format}
\bibliography{sample-base}

\end{document}